\documentclass[letterpaper, 10 pt, conference]{ieeeconf}  

\IEEEoverridecommandlockouts                              

\usepackage{amsthm}   
\theoremstyle{definition}
\newtheorem{problem}{Problem}
\usepackage{hyperref}       
\usepackage{url}            
\usepackage{booktabs}       
\usepackage{amsfonts}       
\usepackage{array}
\newcolumntype{L}[1]{>{\centering\arraybackslash}m{#1}}
\renewcommand{\tabcolsep}{3pt}
\usepackage{multirow}
\usepackage{threeparttable}
\usepackage{caption}
\usepackage{subcaption}
\usepackage{graphicx}
\usepackage{float}
\usepackage{amssymb}   
\usepackage{amsmath}

\usepackage[ruled, linesnumbered, vlined]{algorithm2e}  
\SetArgSty{textnormal}
\newcommand{\llIf}[2]{{\let\par\relax\lIf{#1}{#2}}}
\newcommand{\llElse}[1]{{\let\par\relax\lElse{#1}}}
\newcommand{\forcond}{}
\usepackage{cite}

\usepackage{color,soul}

\title{\LARGE \bf
Game Theoretic Decision Making by Actively Learning Human Intentions Applied on Autonomous Driving
}

\author{Siyu Dai, Sangjae Bae, and David Isele
\thanks{*Authors are with Honda Research Institute, USA}
}

\begin{document}

\maketitle
\thispagestyle{empty}
\pagestyle{empty}

\begin{abstract}

The ability to estimate human intentions and interact with human drivers intelligently is crucial for autonomous vehicles to successfully achieve their objectives. In this paper, we propose a game theoretic planning algorithm that models human opponents with an iterative reasoning framework and estimates human latent cognitive states through probabilistic inference and active learning. By modeling the interaction as a partially observable Markov decision process with adaptive state and action spaces, our algorithm is able to accomplish real-time lane changing tasks in a realistic driving simulator. We compare our algorithm's lane changing performance in dense traffic with a state-of-the-art autonomous lane changing algorithm to show the advantage of iterative reasoning and active learning in terms of avoiding overly conservative behaviors and achieving the driving objective successfully.

\end{abstract}

\section{INTRODUCTION}

Interacting with human drivers safely and efficiently is a major challenge faced by autonomous vehicles (AVs). Human driving behaviors can be very diverse, and are influenced by many factors including the driver's personality and the traffic scenario. Obtaining an accurate estimate of the other agents' behavior is crucial to safe interactions in dense traffic. Learning-based approaches for modeling human driving behaviors~\cite{bhattacharyya2019simulating,bouton2020reinforcement,naumann2020analyzing} have achieved impressive results in recent literature, but inevitably face the demand for a large amount of training data and produce trained models that are nontrivial to interpret. Methods based on probabilistic inference~\cite{sekizawa2007modeling, agamennoni2011bayesian,isele2019interactive} provide more interpretable models,
but interactions often requires pre-defining the states of the models. Game theory 
allows us to capture the mutual influence between the traffic agents while modeling human-like interactions, and making fewer restrictive assumptions about state. However, existing work on game-theoretic AV planning often assume agents to be perfectly rational utility maximizers that operate under equilibrium strategies~\cite{dreves2018generalized, yoo2020game}, whereas human agents are known to deviate from equilibrium behaviors due to cognitive limitations~\cite{wright2014level}.

\begin{figure}
  \centering
  \includegraphics[width=0.75\linewidth]{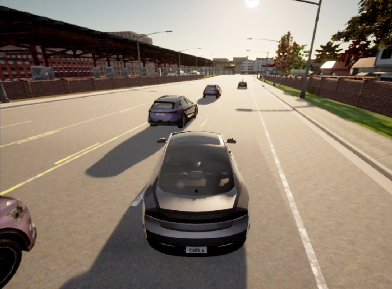}
\caption{\small A visualization of our algorithm running in the CARLA simulation environment. The ego car (Right) negotiates with other traffic to merge into a lane of dense on the left. }
\label{fig:carla}
\vspace{-\baselineskip}
\end{figure}

In this paper, we propose an approach that combines game theory and probabilistic inference in order to reason about human drivers' behavior models, and achieve the AV's objective through successful interactions. In order to take human drivers' diverse driving styles and cognitive levels into account, we utilize the quantal level-$k$ reasoning theory~\cite{stahl1994experimental} and model human's distinct behaviors with \emph{intelligence levels} and \emph{rationality coefficients}. Intelligence levels describe how sophisticated an agent believes its opponents are when reacting to their actions, whereas rationality coefficients describe how close an agent is to a perfect maximizer. We model the interactions between an AV with human drivers as a partially observable Markov decision process (POMDP) and apply probabilistic inference to reason about their intelligence levels and degrees of rationality. Instead of passively adapting to humans' behaviors, our approach actively learns the humans' latent states in order to plan for actions that optimize the AV's own objective. Although similar ideas have previously been discussed in the context of abstract grid-based driving simulators~\cite{tian2021anytime}, our proposed approach can adapt to diverse scenarios with multiple opponent agents and we demonstrate its performance in CARLA~\cite{Dosovitskiy17} (Fig.~\ref{fig:carla}), a more realistic driving simulator. In addition, since our framework employs a standard POMDP formulation, it can be intuitively adapted to many other human-robot interaction applications beyond autonomous driving.

\section{RELATED WORK}

Game theory has been widely applied in autonomous driving for handling vehicle interactions. Stackelberg equilibrium~\cite{von2010market} and Nash equilibrium~\cite{vives1990nash} are commonly used for modeling traffic scenarios. However, when modeled as pure \emph{followers} in Stackelberg games~\cite{fisac2019hierarchical, zhang2019game, hang2020human, yoo2020game}, human drivers are assumed to be fully accommodating,  rather than being treated as agents that may try to influence others. Additionally, human drivers are required to accurately predict AVs' behaviors, which is hardly realistic in practical scenarios. On the other hand, when adopting the Nash equilibrium solution~\cite{dreves2018generalized, liniger2019noncooperative, wang2021game}, players are commonly assumed to have unlimited computational resources and respond optimally to the others, whereas humans are known to have cognitive limitations and don't act with perfect rationality. Williams et al. \cite{williams2018best} proposed Iterative Best Response (IBR) and formulated vehicle interactions as differential games that can be solved approximately using optimization, but their approach requires knowledge of the other vehicles’ dynamics, objectives, and current state. Zhan et al. \cite{Zhan-RSS-21} applies game theoretic techniques to actively gather information during the interaction, but focused on observing occluded opponents instead of predicting and reacting to opponents’ behavior. Li et al. \cite{li2019decision} and Tian et al. \cite{tian2021anytime} applied hierarchical cognitive frameworks based on level-$k$ reasoning~\cite{stahl1994experimental} to explicitly model the cognitive limitations and irrationality of human agents, but limited the evaluation of their models to abstract grid-based driving simulators. In contrast, our work adapts game-theoretic formulations to realistic driving scenarios, and we demonstrate the real-time planning performance of our approach in lane-changing scenarios in dense traffic.





\section{PRELIMINARIES}

\subsection{Quantal Level-$k$ Reasoning} 

Quantal level-$k$ reasoning~\cite{breitmoser2014beliefs} is a game theoretic modeling technique that combines the ideas of quantal response equilibrium (QRE) models~\cite{mckelvey1995quantal} and level-$k$ reasoning~\cite{stahl1994experimental}. QRE models the imperfect maximization of expected payoffs using stochastic choices, hence it is a more realistic model for human players than the sharp best responses in standard game theory. Level-k reasoning theory additionally models how players form conjectures about their co-players’ strategies and assumes players choose actions according to their belief. In the quantal level-$k$ (ql-$k$) reasoning model, players are divided into $k$ intelligence levels and they react to opponent actions by assuming the opponents are at a lower intelligent level than themselves. A ql-0 agent is assumed to be non-strategic, and a ql-$k$ agent, $k \in \mathbb{N}^+$, believes that its opponent is a ql-$(k-1)$ agent and chooses actions according to the QRE model. The stochastic behaviors of each agent is modeled by a rationality coefficient $\lambda \in \mathbb{R}^+$. Let $Q^i(s, a^i|a^{-i})$ denote the expected total reward agent $i$ achieves when executing action $a^i$ at state $s$ against an action $a^{-i}$ from its opponent $-i$, then the policy of a ql-$k$ agent with rationality $\lambda$ can be expressed as:

\begin{equation}
\label{equ:level_k_policy}
    \pi ^i (a^i|s, a^{-i}) = \frac{\exp (\lambda Q^i(s, a^i|a^{-i}))}{\sum_{a' \in \mathcal{A}^i} \exp (\lambda Q^i(s, a'|a^{-i}))},
\end{equation}

\noindent where $\mathcal{A}^i$ denotes the action space of agent $i$.



\subsection{Monte Carlo Tree Search (MCTS)}

MCTS~\cite{chaslot2008monte} is a heuristic search approach widely used in problems where the exact solutions are difficult or impossible to solve for. It breaks the curse of dimensionality by sampling state transitions instead of considering all possible state transitions when estimating the potential long-term reward. It converges to the optimal policy when exploration is controlled appropriately~\cite{silver2010monte}. During rollouts, MCTS selects the next node to expand greedily according to the UCT algorithm~\cite{kocsis2006bandit} which augments the action reward by an exploration bonus. It continues to conduct rollouts until timeout and returns the current best action. Similar to Silver et al. \cite{silver2010monte}, we apply MCTS to solve POMDPs, but we adapt it to work with multi-player games and incorporate an active learning component.

\section{PROBLEM STATEMENT}

We model the decision making problem faced by a robotic agent when interacting with humans as a POMDP. $\mathbf{s} = [\mathbf{x}, \mathbf{\theta}]$ represents a state in the POMDP, where $\mathbf{x}$ denotes the physical states (including $\mathbf{x}^e$ for ego agent and $\mathbf{x}^i$, $i = 1, 2, \ldots, n$ for $n$ human agents) and $\mathbf{\theta}$ denotes the internal states that dominate the opponents' behavior.
We assume all physical states are fully observable and all opponents' internal states are hidden, and denote the observations as $\mathbf{o}$.
We denote the action space of the ego agent as $\mathcal{A}^{\mathcal{E}}$, and assume all human agents have the same action space $\mathcal{A}^{\mathcal{H}}$. We assume the opponents hidden state $\mathbf{\theta}$ doesn't change with $t$ during the decision making process. The ego agent maintains a belief on the probability distribution over states $\mathbf{b}(\mathbf{s})$. Then the POMDP problem can be written as follows:

\begin{problem}
\label{problem_def}
\begin{equation*}
  \begin{aligned}
& \underset{\pi}{\textrm{maximize}} & & V(\pi) = \mathbb{E}\left[\sum_{t=0}^{\infty} \gamma^t r(\mathbf{b}_t, \mathbf{a}^e_t)|\mathbf{a}^e_t \sim \pi \right] \\
& \textrm{subject to} & & \mathbf{x}_{t+1} = f(\mathbf{x}_t, \mathbf{a}^e_t, \mathbf{a}^1_t, \ldots, \mathbf{a}^n_t) \\
& & & \mathbf{a}^e_t \in \mathcal{A}^{\mathcal{E}},~ \mathbf{a}^i_t \in \mathcal{A}^{\mathcal{H}}, ~~ i = 1, 2, \ldots, n \\
& & & \mathbf{o} = \mathbf{x},~ \mathbf{s} = [\mathbf{x}, \mathbf{\theta}]  \\
& & & \mathbf{b}_{t+1}(\mathbf{s}_{t+1}) = \rho (\mathbf{b}_t(\mathbf{s}_t), \mathbf{o}_{t+1}), \\
\end{aligned}  
\end{equation*}
\end{problem}

\noindent where $r$ is the reward function, $\gamma$ is the discount factor, $f$ is a known dynamics function that governs the transition of physical states and $\rho$ is the belief update function.

\section{APPROACH}

To solve Problem~\ref{problem_def}, we propose to model human opponents as ql-$k$ agents and pre-compute their policies offline. We then let the hidden state $\mathbf{\theta} = [k^i, \lambda ^i]$, $i = 1, 2, \ldots, n$ and solve the POMDP online with MCTS by propagating beliefs according to the pre-computed policies.

\subsection{Quantal Level-$k$ Policies}


In order to achieve real-time belief update during online planning, we pre-compute the $Q$ value function for perfectly rational agents at each level $k$ to store in memory, and then compute the quantal policies using Equation~\ref{equ:level_k_policy} according to the ego agent's belief on $k^i$ and $\lambda ^i$ during online belief propagation. We assume the non-strategic level-$0$ policy is given, and a perfectly rational level-$k$ agent maximizes the objective $V^{i, k}_{\pi ^ {i, k}}(\mathbf{s}_0) = \mathbb{E}_{\pi ^ {-i, k-1}} [\sum_{t=0}^{\infty} \gamma^t r(\mathbf{s}_t, \mathbf{a}^{i, k}_t) | \mathbf{a}^{i, k}_t \sim \pi ^ {i, k}]$. Note that we assume all humans follow the ql-$k$ policies and only differ in terms of the internal states $\mathbf{\theta}$, hence during the value function pre-computation, we only consider a two-player game with one human and one ego agent. $V^{i, k}$ can be computed using value iteration~\cite{bellman1957markovian} when the state space is discrete, and fitted value iteration~\cite{pmlr-v139-lutter21a} when the state space is continuous. With the value function $V^{i, k}$, we can then compute the $Q$ function using the following equation:

\begin{equation}
    \label{equ:Q}
    Q^{i, k}(\mathbf{s}_t, \mathbf{a}^{i, k}_t) = \mathbb{E}_{\pi ^ {-i, k-1}} [r(\mathbf{s}_t, \mathbf{a}^{i, k}_t) + \gamma V^{i, k}(\mathbf{s}_{t+1})].
\end{equation}

\noindent In contrast to the quantal level-$k$ dynamic programming approach proposed by Tian et al.~\cite{tian2021anytime} which directly computes the level-$k$ policies during offline training, we only pre-compute the $Q$ functions offline and compute ql-$k$ policies during online planning. This allows us to save memory space because we don't need to store the level-$k$ policies for every possible $\lambda$ value. In addition, it also makes it more intuitive to extend our framework to incorporate the entire continuous set of $\lambda$ values instead of pre-specifying a few discrete values.

\subsection{Game Theoretic Planning with MCTS}


During online planning, we construct a search tree which stores the current observation and belief with the root node as $root.\mathbf{x}$ and $root.\mathbf{b}$ respectively. In order to avoid the exponential increase of the tree expansion breadth as the number of opponents increases, we select up to two opponents during tree search according to their relative positions to the ego vehicle. We first identify the nearest two human agents, and if both the nearest agents are in front of or behind the ego agent, we only consider the nearest one as the opponent. Otherwise, both the nearest agents are considered as opponents. All other agents in the scene are considered as obstacles.

\begin{algorithm}[!t]
\small
 \caption{Game Theoretic Planning}
 \label{algorithm:full}
 \DontPrintSemicolon
 \KwIn{\\ $\mathcal{E}: \text{environment}$
 \\ $n$: \text{total number of other vehicles}
 \\ $H$: \text{planning horizon}
 \\ $T$: \text{time allowance for MCTS}
 \\ $\gamma$: \text{discount factor}
 \\ $\phi$: \text{information gain reward coefficient}
 \\ $\mathbf{b}_0$: \text{initial belief}
 \\ $Q^k$: \text{precomputed} $Q$ \text{function,} $k = 1, \ldots, k_{max}$
}
 
 $\textbf{o} \leftarrow $ take initial observation \\
 $oppo\_id \leftarrow $ SelectOpponent($\textbf{o}$) \\
 $Tree \leftarrow $ ConstructTree($\textbf{o}, \mathbf{b}_0, oppo\_id$)\\
 \While{\textbf{not} success}
 {
 $best\_action \leftarrow Tree.\text{search}(H, T)$ \\
 Execute $best\_action$ and take observation $\textbf{o}$ in $\mathcal{E}$ \\
 $oppo\_id \leftarrow $ SelectOpponent($\textbf{o}$) \\
 $Tree.$UpdateRoot($\textbf{o}$) \\
 $Tree.$UpdateOppo($oppo\_id$) \\
 }
 \;
 \SetKwFunction{FSearch}{Tree.search}
 \SetKwProg{Fn}{Function}{:}{}
 \Fn{\FSearch{$H, T$}}{
 \lWhile{$time\_elapsed < T$}{
 $Tree$.rollout($root, H$)
 }
 $best\_action \leftarrow$ argmax$_\mathbf{a}~root.values(\mathbf{a})$ \\
 \Return $best\_action$
 }
  \;
 \SetKwFunction{FRollout}{Tree.rollout}
 \Fn{\FRollout{$node, H$}}{
 \llIf{$node.depth = H$}{\Return 0} \\
 \eIf{len($node.unsampled\_as) >$ 0}{
 $\mathbf{a}^e \leftarrow $ random action from $node.unsampled\_as$
 }{
 $\mathbf{a}^e \leftarrow $argmax$_\mathbf{a} ~node.values(\mathbf{a}) + c \sqrt{\frac{\log node.N}{node.N_a(\mathbf{a})}} $
 }
 \For{\forcond $i$ in $n$}{
 \eIf{$i$ in $oppo\_id$}{
 $\pi^o \leftarrow$ ComputePolicy($node.\mathbf{b}[i], Q^k$) \\
 Sample $\mathbf{a}^o[i] \sim \pi^o$
 }{$\mathbf{a}^o[i] \leftarrow default\_action$}}
 \llIf{$(\mathbf{a}^e, \mathbf{a}^o)$ \textbf{not in $node.children$}}{expand($\mathbf{a}^e, \mathbf{a}^o$)} \\
 $child \leftarrow node.children(\mathbf{a}^e, \mathbf{a}^o)$ \\
 $R = child.r + \gamma * Tree.$rollout$(child, H)$ \\
 $node.N ~+= 1$ \\
 $node.N_a(\mathbf{a}^e) ~+= 1$ \\
 $node.values(\mathbf{a}^e) ~+= \frac{R - node.values(\mathbf{a}^e)}{node.N_a(\mathbf{a}^e)}$ \\
 \Return $R$
 }
   \;
 \SetKwFunction{FExpand}{node.expand}
 \Fn{\FExpand{$\mathbf{a}^e, \mathbf{a}^o$}}{
 $child.\mathbf{x}, r \leftarrow $ Simulate$(node.\mathbf{x}, \mathbf{a}^e, \mathbf{a}^o)$ \\
 $child.\mathbf{b} \leftarrow $ BeliefUpdate$(node.\mathbf{x}, node.\mathbf{b}, child.\mathbf{x})$ \\
 $info\_gain = \sum_{i \in oppo\_id} (\mathcal{H}(node.\mathbf{b}) - \mathcal{H}(child.\mathbf{b})) $ \\
 $child.r = r + \phi * info\_gain$ \\
 $node.children(\mathbf{a}^e, \mathbf{a}^o) \leftarrow child$
 }
\end{algorithm}

Algorithm~\ref{algorithm:full} describes the game theoretic planning procedure. It relies on a tree structure that conducts rollouts to a depth that equals to the planning horizon $H$ until timeout and returns the current best action based on the rewards of the rollouts. The tree consists of nodes that store the state $node.\mathbf{x}$, a list of children indexed by the ego action $\mathbf{a}^e$ and opponent actions $\mathbf{a}^o$, belief on opponents' hidden states $node.\mathbf{b}$, and the values of different ego actions $node.values$. During rollout, ego actions that have never been sampled ($node.unsampled\_as$) are prioritized, and if there are no more unsampled actions, an action with the highest UCT score~\cite{kocsis2006bandit} will be selected, which considers both its value and the number of times it has been sampled (line 18-21). Human agents that are not viewed as opponents are assumed to take the no-op action, i.e. maintaining constant velocity. Opponent agents' ql-$k$ policies are computed using Equation~\ref{equ:level_k_policy} with pre-computed $Q^k$ value functions, and their actions are sampled from the policy distribution weighted by the belief state $node.\mathbf{b}$. Each node keeps track of the total number of times it has been sampled ($node.N$) and the number of times each action has been sampled ($node.N_a$). The value of each action is computed by averaging the reward each rollout has achieved (line 33). 

During node expansion, the state of the child node $child.\mathbf{x}$ and the reward $r$ are estimated by the simulator, and the belief state of the child node $child.\mathbf{b}$ is updated following the standard POMDP belief update procedure based on the opponent actions $\mathbf{a}^o$ that led to the expansion of this child node (line 37-38 in Algorithm~\ref{algorithm:full}). In order to facilitate active learning of the human latent states, we augment the reward with an information gain reward that measures the reduction in the total entropy of the belief states on each opponent agent. We use an information gain reward coefficient $\phi$ to control the weight of this information gain reward during rollouts (line 39-40).

\begin{figure*}
  \centering
  \includegraphics[width=0.84\linewidth]{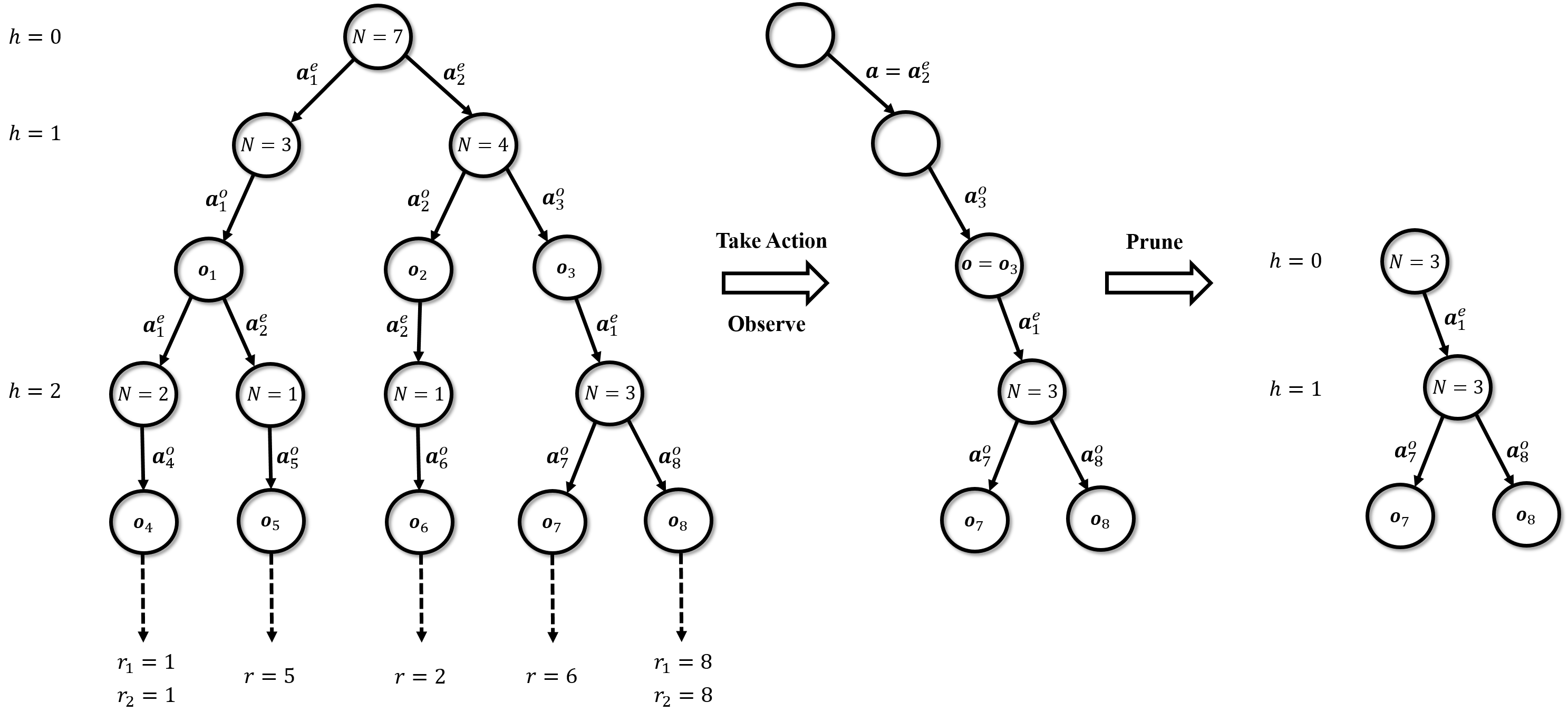}
\caption{\small An illustrative diagram of one search step of our algorithm with planning horizon $H = 2$ and two available ego vehicle actions. The leftmost graph shows the search tree after conducting 7 rollouts within the time allowance $T$ (line 12 in Algorithm~\ref{algorithm:full}), where $h$ denotes the depth, $N$ denotes the total number of times a node has been expanded, and $r$ denotes the total reward of a rollout trajectory. $a^e_1$ and $a^e_2$ represent the two available ego actions, and $a^o_i$ includes the actions of all other vehicles. After the algorithm decides that $a^e_2$ is the best action to take at the current time step, it executes $a^e_2$ and receives a real observation from the environment. The algorithm identifies the real observation $\mathbf{o} = \mathbf{o}_3$, hence it prunes the section of the search tree that is no longer possible and reassigns the root node.}
\label{fig:diagram}
\vspace{-\baselineskip}
\end{figure*}

After a best action is returned by the search tree, ego agent executes it and observes the environment's physical state $\mathbf{o}$. When planning for the next action, we reevaluate which human agents should be considered as opponents, and prune sections of the tree that are no longer possible by reassigning the current observation as the new root node. This search procedure is repeated until the goal is achieved. An example illustration of our algorithm with planning horizon $H = 2$ and two available ego actions is shown in Fig.~\ref{fig:diagram}.

Our game theoretic planning algorithm differs from the one proposed by Tian et al.~\cite{tian2021anytime} in three major aspects. First, instead of pre-computing and storing in memory the entire transition matrix of the POMDP, we takes advantage of the discretized state and action space and the known dynamics function to conduct online tree expansion and rollouts. The approach in \cite{tian2021anytime} is constrained within abstract grid-world applications due to its high memory consumption and requirement of re-training the level-$k$ policies and transition matrices when opponents have unseen rationality coefficient values. In contrast, our approach is adaptable to various scenarios through state-space mapping and can still maintain real-time planning performance through MCTS. Second, the ego vehicle in \cite{tian2021anytime} is restricted to follow a ql-$k$ policy and the values of tree nodes are estimated using pre-computed ql-$k$ value functions, whereas our ego agent is able to plan for the best action according to the true value without being restricted by the ql-$k$ policy. Additionally, the approach in \cite{tian2021anytime} can only be applied in scenarios with one other agent, whereas our approach can be adapted to scenarios with an arbitrary number of human agents in the scene.

\section{EMPIRICAL EVALUATION}

We evaluate the performance of our proposed approach on a lane changing task in the CARLA~\cite{Dosovitskiy17} simulator. All experiments are conducted on a 16-core Intel i9 2.4 GHz laptop with 32 GB RAM and one GeForce RTX 2060 GPU.

\subsection{Experiment Setup}

In order to achieve real-time planning performance, we use a small discretized state and action space instead of the real CARLA simulator for MCTS rollouts and pre-training ql-$k$ policies. We assume $k_{max} = 2$ based on the findings on human intelligence levels in the literature ~\cite{costa2006cognition,costa2009comparing}, and set the planning horizon depth $H = 4$ which corresponds to a $4T$-second look ahead. The agent receives a negative reward at every step where it's not on the target lane and a large negative reward for collision. The non-strategic level-$0$ policy is computed by assuming all other agents in the scene are static obstacles. The $Q$ functions for ql-$k$ policies are pre-trained by first obtaining the value function through value iteration assuming opponent agents are ql-$(k-1)$ agents, and then computing $Q$ using Equation~\ref{equ:Q}. The game theoretic planner will publish high-level target trajectories, and a low-level PID controller is used to track the target trajectory. To ensure adaptability in various scenarios, we use a $40~m$ long two-lane wide grid centered around the ego vehicle as the state space, and conduct real-time state mapping between the grid state space and the true physical state space during online planning. The size of the grid system also means that only vehicles that are on the ego lane and target lane with a distance of less than $20~m$ to the ego vehicle will be considered by the planner. The ego agent has five actions: ``accelerate", ``decelerate", ``maintain velocity" (no-op), ``indicate lane change intention" and ``lane change". The ``accelerate" action causes the ego vehicle to increase its velocity by $1~m/s$ and the ``decelerate" action causes to decrease velocity by $1~m/s$. The ego agent starts at rest, and traffic is initialized at $3~m/s$. The ``indicate lane change intention" action will trigger a target trajectory that moves the ego vehicle $0.5~m$ towards the target lane while maintaining the same longitudinal speed, whereas the ``lane change" action will publish a target trajectory that brings the ego vehicle to the center of the target lane. To simplify the scenario, we encode only three actions for the human agents: accelerate, decelerate and maintain velocity (no-op).

The scenarios we tested on are in the \verb Town03 ~environment in CARLA with an illegal bicycle and a stopped vehicle $100~m$ ahead of the ego vehicle's starting pose, forcing a lane change. We evaluate our planner in three non-cooperative game scenarios with one, two and six driving vehicles in the target lane respectively, where the driving vehicles follow the Intelligent Driver Model (IDM)~\cite{treiber2000congested}. The task goal is to reach within $0.5~m$ in the lateral direction from the center of the target lane before reaching the stopped vehicle in the ego lane. The driving vehicles are placed in the scene to simulate human drivers, and their controller tracks a randomly generated velocity while maintaining a distance of $5-10~m$ with the front vehicle. We program the other vehicles to not follow ql-$k$ policies in order to show the generalization behavior of our planner to unseen behaviors. We simulate the behaviors of aggressive drivers and yielding drivers by allowing these driving vehicles to yield when the ego vehicle is close in half of the tests for each set of experiments. We compare our planner's lane changing performance with NNMPC~\cite{bae2020cooperation}, a state-of-the-art autonomous lane changing planner that uses Recurrent Neural Networks (RNNs) for predicting other vehicles' behaviors and guiding the Model Predictive Controller (MPC). We also conduct an ablation study by comparing our approach with a version without the information gain reward (Algorithm~\ref{algorithm:full} line 40) in order to show the effect of active learning.

\subsection{Evaluation on Belief Accuracy}

\begin{table}
\caption{Evaluation on Belief Accuracy}
\label{tab:belief}
\small\centering
\begin{tabular}{L{1.7cm}|L{0.9cm}|L{2cm}|L{2cm}}
\toprule
\# Opponents & $T$ (s) & With Info Gain & W/O Info Gain \\ \hline
\multirow{3}{1.7cm}{\centering One} & 0.5 & 87.5\% & 81\% \\
 & 1 & 85\% & 80.5\% \\ 
 & 2 & 85.5\% & 80.5\% \\ \hline
 \multirow{3}{1.7cm}{\centering Two} & 0.5 & 82.5\% & 79.8\% \\
 & 1 & 83.9\% & 79.3\% \\ 
 & 2 & 83.8\% & 80.9\% \\ 
 \bottomrule
\end{tabular}
\vspace{-\baselineskip}
\end{table}

We first evaluate whether our game theoretic planner can accurately estimate the intelligence level of opponent agents. Since up to two agents can be considered as opponents at a time, we experiment with two scenarios: one opponent in the scene and two opponents in the scene. We compare the performance using time allowances $T = 0.5,~T = 1$ and $T = 2$, and conduct an ablation study by comparing our full model against a model without the information gain. In each scenario, we let the opponent follow a ql-$k$ policy for all $k = 1, \ldots, k_{max}$ (or all combinations of $k$'s when there are two opponents) and run 100 simulations with rationality coefficients $\lambda$ randomly sampled from [1, 3, 5] for each $k$. We view a belief state as accurate if it assigns a larger than 0.5 probability for the correct intelligence level. We average the accuracy for all $k$ combinations in each set of experiments and show the results in Table~\ref{tab:belief}. We see that with information gain reward the average performance is better. 
This shows that active learning improves the ego agent's estimation accuracy on the opponents' latent states. Even though accurately estimating belief is more challenging when there are two opponents in the scene, our approach is still able to achieve an accuracy of above 80\%.

\subsection{Evaluation on Lane Changing Performance}

Table~\ref{tab:lane_change} compares the autonomous lane changing performance of our game theoretic planner, its version without information gain reward, and NNMPC, where each entry is the average of 50 test runs in the CARLA simulator. In the experiments in this section, the time allowance in our algorithm is set to $T=1~s$. From the results in Table~\ref{tab:lane_change} and the qualitative performance in the simulation videos we observed that NNMPC is a very conservative planner that will sacrifice the time to merge to obtain low collision rate. It takes more than three times as much time to change lane compared to our approach in all the scenarios we tested, and we often observe it stopping to wait for all vehicles to pass before merging in, which causes it to have a higher time-out rate in dense traffic (i.e. scenarios with 6 other vehicles). This is because even though NNMPC is able to predict other agents' paths and plan for a safe trajectory to merge, it lacks the ability to actively interact with the other agents and signal a lane-changing intention. On the contrary, our approach leverages game theory to generate interactive behaviors that help the ego agent achieve its goal as soon as possible. For example, we observe that when our planner meets aggressive opponents, it will slow down to let the opponent go first and then merge in after, whereas NNMPC in the same situation will continue going forward with the same speed and reaches a dead-lock situation with the opponent vehicle. Similarly, when our planner signaled the intention to merge and found that the opponent vehicle is yielding, it then accelerated to a safer location to accomplish the lane change. NNMPC never makes the attempt to change lane as long as the opponent vehicles are close and therefore is not always able to take advantage of potentially yielding opponents.

\begin{table}
\caption{Evaluation on Lane Changing Performance}
\label{tab:lane_change}
\small\centering
\begin{threeparttable}
\begin{tabular}{L{1.1cm}|L{2.3cm}|L{1.4cm}|L{1.4cm}|L{1.2cm}}
\toprule
Scenario & Metric\tnote{1} & Ours with Info Gain & Ours w/o Info Gain & NNMPC \\ \hline
\multirow{3}{1.1cm}{\centering 1 other vehicle} & time to merge (s) & 5.52 & 7.02 & 20.08 \\
 & collision rate & 2\% & 6\% & 0\% \\ 
 & timeout rate & 0\% & 4\% & 0\% \\ \hline
 \multirow{3}{1.1cm}{\centering 2 other vehicles} & time to merge (s) & 7.5 & 8.96 & 24.69 \\
 & collision rate & 4\% & 12\% & 2\% \\ 
 & timeout rate & 2\% & 8\% & 0\% \\ \hline
 \multirow{3}{1.1cm}{\centering 6 other vehicles} & time to merge (s) & 10.2 & 12.6 & 31.5 \\
 & collision rate & 8\% & 12\% & 4\% \\ 
 & timeout rate & 6\% & 10\% & 16\% \\ 
 \bottomrule
\end{tabular}
\begin{tablenotes}
\footnotesize
 \item[1] Time to merge refers to the time it takes the ego vehicle to reach within $0.5~m$ from the center of the target lane, and timeout includes failure to merge when reached within $10~m$ from the stopped vehicle or having stopped for more than $15~s$ in the ego lane.
\end{tablenotes}
\end{threeparttable}
\vspace{-\baselineskip}
\end{table}

Another finding from Table~\ref{tab:lane_change} is that the time to merge and failure rate are both higher when our approach is used without the information gain reward, which is likely caused by the less accurate estimation of opponent intelligence level. In general, our approach has a relatively higher collision rate compared to NNMPC. This issue can potentially be solved by adding chance constraints~\cite{ono2008iterative, dai2019chance} to our planner or by using a collision-aware low-level controller to execute the high-level target trajectories.

\section{CONCLUSIONS}

In this paper, we propose a game theoretic autonomous driving algorithm that can actively reason about human drivers' latent cognitive states in real time and achieve its driving objective through successful interactions with other vehicles. By modeling human drivers as level-$k$ reasoning agents and modeling the interaction as a POMDP, our algorithm is able to estimate humans' latent cognitive states through probabilistic inference. By augmenting the reward function with an information gain reward, our agent can actively take actions to encourage the accurate estimation of latent states instead of passively reacting to human behaviors. We apply adaptive state and action spaces in the POMDP model and utilize MCTS to achieve real-time planning performance in lane changing scenarios in the CARLA driving simulator. Experiment results comparing our algorithm with another state-of-the-art autonomous lane changing algorithm show that our approach is able to avoid overly conservative behaviors and succeed more frequently in mandatory lane changing scenarios in dense traffic.










\bibliographystyle{IEEEtran}
\bibliography{root}

\begin{thebibliography}{10}
\providecommand{\url}[1]{#1}
\csname url@rmstyle\endcsname
\providecommand{\newblock}{\relax}
\providecommand{\bibinfo}[2]{#2}
\providecommand\BIBentrySTDinterwordspacing{\spaceskip=0pt\relax}
\providecommand\BIBentryALTinterwordstretchfactor{4}
\providecommand\BIBentryALTinterwordspacing{\spaceskip=\fontdimen2\font plus
\BIBentryALTinterwordstretchfactor\fontdimen3\font minus
  \fontdimen4\font\relax}
\providecommand\BIBforeignlanguage[2]{{%
\expandafter\ifx\csname l@#1\endcsname\relax
\typeout{** WARNING: IEEEtran.bst: No hyphenation pattern has been}%
\typeout{** loaded for the language `#1'. Using the pattern for}%
\typeout{** the default language instead.}%
\else
\language=\csname l@#1\endcsname
\fi
#2}}

\bibitem{bhattacharyya2019simulating}
R.~P. Bhattacharyya, D.~J. Phillips, C.~Liu, J.~K. Gupta, K.~Driggs-Campbell,
  and M.~J. Kochenderfer, ``Simulating emergent properties of human driving
  behavior using multi-agent reward augmented imitation learning,'' in
  \emph{2019 International Conference on Robotics and Automation (ICRA)}.\hskip
  1em plus 0.5em minus 0.4em\relax IEEE, 2019, pp. 789--795.

\bibitem{bouton2020reinforcement}
M.~Bouton, A.~Nakhaei, D.~Isele, K.~Fujimura, and M.~J. Kochenderfer,
  ``Reinforcement learning with iterative reasoning for merging in dense
  traffic,'' in \emph{2020 IEEE 23rd International Conference on Intelligent
  Transportation Systems (ITSC)}.\hskip 1em plus 0.5em minus 0.4em\relax IEEE,
  2020, pp. 1--6.

\bibitem{naumann2020analyzing}
M.~Naumann, L.~Sun, W.~Zhan, and M.~Tomizuka, ``Analyzing the suitability of
  cost functions for explaining and imitating human driving behavior based on
  inverse reinforcement learning,'' in \emph{2020 IEEE International Conference
  on Robotics and Automation (ICRA)}.\hskip 1em plus 0.5em minus 0.4em\relax
  IEEE, 2020, pp. 5481--5487.

\bibitem{sekizawa2007modeling}
S.~Sekizawa, S.~Inagaki, T.~Suzuki, S.~Hayakawa, N.~Tsuchida, T.~Tsuda, and
  H.~Fujinami, ``Modeling and recognition of driving behavior based on
  stochastic switched arx model,'' \emph{IEEE Transactions on Intelligent
  Transportation Systems}, vol.~8, no.~4, pp. 593--606, 2007.

\bibitem{agamennoni2011bayesian}
G.~Agamennoni, J.~I. Nieto, and E.~M. Nebot, ``A bayesian approach for driving
  behavior inference,'' in \emph{2011 IEEE Intelligent Vehicles Symposium
  (IV)}.\hskip 1em plus 0.5em minus 0.4em\relax IEEE, 2011, pp. 595--600.

\bibitem{isele2019interactive}
D.~Isele, ``Interactive decision making for autonomous vehicles in dense
  traffic,'' in \emph{2019 IEEE Intelligent Transportation Systems Conference
  (ITSC)}.\hskip 1em plus 0.5em minus 0.4em\relax IEEE, 2019, pp. 3981--3986.

\bibitem{dreves2018generalized}
A.~Dreves and M.~Gerdts, ``A generalized nash equilibrium approach for optimal
  control problems of autonomous cars,'' \emph{Optimal Control Applications and
  Methods}, vol.~39, no.~1, pp. 326--342, 2018.

\bibitem{yoo2020game}
J.~Yoo and R.~Langari, ``A game-theoretic model of human driving and
  application to discretionary lane-changes,'' \emph{arXiv preprint
  arXiv:2003.09783}, 2020.

\bibitem{wright2014level}
J.~R. Wright and K.~Leyton-Brown, ``Level-0 meta-models for predicting human
  behavior in games,'' in \emph{Proceedings of the fifteenth ACM conference on
  Economics and computation}, 2014, pp. 857--874.

\bibitem{stahl1994experimental}
D.~O. Stahl~II and P.~W. Wilson, ``Experimental evidence on players' models of
  other players,'' \emph{Journal of economic behavior \& organization},
  vol.~25, no.~3, pp. 309--327, 1994.

\bibitem{tian2021anytime}
R.~Tian, L.~Sun, M.~Tomizuka, and D.~Isele, ``Anytime game-theoretic planning
  with active reasoning about human's latent states for human-centered
  robots,'' in \emph{2021 International Conference on Robotics and Automation
  (ICRA)}.\hskip 1em plus 0.5em minus 0.4em\relax IEEE, 2021.

\bibitem{Dosovitskiy17}
A.~Dosovitskiy, G.~Ros, F.~Codevilla, A.~Lopez, and V.~Koltun, ``{CARLA}: {An}
  open urban driving simulator,'' in \emph{Proceedings of the 1st Annual
  Conference on Robot Learning}, 2017, pp. 1--16.

\bibitem{von2010market}
H.~Von~Stackelberg, \emph{Market structure and equilibrium}.\hskip 1em plus
  0.5em minus 0.4em\relax Springer Science \& Business Media, 2010.

\bibitem{vives1990nash}
X.~Vives, ``Nash equilibrium with strategic complementarities,'' \emph{Journal
  of Mathematical Economics}, vol.~19, no.~3, pp. 305--321, 1990.

\bibitem{fisac2019hierarchical}
J.~F. Fisac, E.~Bronstein, E.~Stefansson, D.~Sadigh, S.~S. Sastry, and A.~D.
  Dragan, ``Hierarchical game-theoretic planning for autonomous vehicles,'' in
  \emph{2019 International Conference on Robotics and Automation (ICRA)}.\hskip
  1em plus 0.5em minus 0.4em\relax IEEE, 2019, pp. 9590--9596.

\bibitem{zhang2019game}
Q.~Zhang, R.~Langari, H.~E. Tseng, D.~Filev, S.~Szwabowski, and S.~Coskun, ``A
  game theoretic model predictive controller with aggressiveness estimation for
  mandatory lane change,'' \emph{IEEE Transactions on Intelligent Vehicles},
  vol.~5, no.~1, pp. 75--89, 2019.

\bibitem{hang2020human}
P.~Hang, C.~Lv, Y.~Xing, C.~Huang, and Z.~Hu, ``Human-like decision making for
  autonomous driving: A noncooperative game theoretic approach,'' \emph{IEEE
  Transactions on Intelligent Transportation Systems}, vol.~22, no.~4, pp.
  2076--2087, 2020.

\bibitem{liniger2019noncooperative}
A.~Liniger and J.~Lygeros, ``A noncooperative game approach to autonomous
  racing,'' \emph{IEEE Transactions on Control Systems Technology}, vol.~28,
  no.~3, pp. 884--897, 2019.

\bibitem{wang2021game}
M.~Wang, Z.~Wang, J.~Talbot, J.~C. Gerdes, and M.~Schwager, ``Game-theoretic
  planning for self-driving cars in multivehicle competitive scenarios,''
  \emph{IEEE Transactions on Robotics}, 2021.

\bibitem{williams2018best}
G.~Williams, B.~Goldfain, P.~Drews, J.~M. Rehg, and E.~A. Theodorou, ``Best
  response model predictive control for agile interactions between autonomous
  ground vehicles,'' in \emph{2018 IEEE International Conference on Robotics
  and Automation (ICRA)}.\hskip 1em plus 0.5em minus 0.4em\relax IEEE, 2018,
  pp. 2403--2410.

\bibitem{Zhan-RSS-21}
Z.~Zhang and J.~F. Fisac, ``{Safe Occlusion-Aware Autonomous Driving via
  Game-Theoretic Active Perception},'' in \emph{Proceedings of Robotics:
  Science and Systems}, Virtual, July 2021.

\bibitem{li2019decision}
S.~Li, N.~Li, A.~Girard, and I.~Kolmanovsky, ``Decision making in dynamic and
  interactive environments based on cognitive hierarchy theory, bayesian
  inference, and predictive control,'' in \emph{2019 IEEE 58th Conference on
  Decision and Control (CDC)}.\hskip 1em plus 0.5em minus 0.4em\relax IEEE,
  2019, pp. 2181--2187.

\bibitem{breitmoser2014beliefs}
Y.~Breitmoser, J.~H. Tan, and D.~J. Zizzo, ``On the beliefs off the path:
  Equilibrium refinement due to quantal response and level-k,'' \emph{Games and
  Economic Behavior}, vol.~86, pp. 102--125, 2014.

\bibitem{mckelvey1995quantal}
R.~D. McKelvey and T.~R. Palfrey, ``Quantal response equilibria for normal form
  games,'' \emph{Games and economic behavior}, vol.~10, no.~1, pp. 6--38, 1995.

\bibitem{chaslot2008monte}
G.~Chaslot, S.~Bakkes, I.~Szita, and P.~Spronck, ``Monte-carlo tree search: A
  new framework for game ai.'' \emph{AIIDE}, vol.~8, pp. 216--217, 2008.

\bibitem{silver2010monte}
D.~Silver and J.~Veness, ``Monte-carlo planning in large pomdps,''
  \emph{Advances in Neural Information Processing Systems}, vol.~23, pp.
  2164--2172, 2010.

\bibitem{kocsis2006bandit}
L.~Kocsis and C.~Szepesv{\'a}ri, ``Bandit based monte-carlo planning,'' in
  \emph{European conference on machine learning}.\hskip 1em plus 0.5em minus
  0.4em\relax Springer, 2006, pp. 282--293.

\bibitem{bellman1957markovian}
R.~Bellman, ``A markovian decision process,'' \emph{Journal of mathematics and
  mechanics}, vol.~6, no.~5, pp. 679--684, 1957.

\bibitem{pmlr-v139-lutter21a}
M.~Lutter, S.~Mannor, J.~Peters, D.~Fox, and A.~Garg, ``Value iteration in
  continuous actions, states and time,'' in \emph{Proceedings of the 38th
  International Conference on Machine Learning}, 2021, pp. 7224--7234.

\bibitem{costa2006cognition}
M.~A. Costa-Gomes and V.~P. Crawford, ``Cognition and behavior in two-person
  guessing games: An experimental study,'' \emph{American economic review},
  vol.~96, no.~5, pp. 1737--1768, 2006.

\bibitem{costa2009comparing}
M.~A. Costa-Gomes, V.~P. Crawford, and N.~Iriberri, ``Comparing models of
  strategic thinking in van huyck, battalio, and beil's coordination games,''
  \emph{Journal of the European Economic Association}, vol.~7, no. 2-3, pp.
  365--376, 2009.

\bibitem{treiber2000congested}
M.~Treiber, A.~Hennecke, and D.~Helbing, ``Congested traffic states in
  empirical observations and microscopic simulations,'' \emph{Physical review
  E}, vol.~62, no.~2, p. 1805, 2000.

\bibitem{bae2020cooperation}
S.~Bae, D.~Saxena, A.~Nakhaei, C.~Choi, K.~Fujimura, and S.~Moura,
  ``Cooperation-aware lane change maneuver in dense traffic based on model
  predictive control with recurrent neural network,'' in \emph{2020 American
  Control Conference (ACC)}.\hskip 1em plus 0.5em minus 0.4em\relax IEEE, 2020,
  pp. 1209--1216.

\bibitem{ono2008iterative}
M.~Ono and B.~C. Williams, ``Iterative risk allocation: A new approach to
  robust model predictive control with a joint chance constraint,'' in
  \emph{2008 47th IEEE Conference on Decision and Control}.\hskip 1em plus
  0.5em minus 0.4em\relax IEEE, 2008, pp. 3427--3432.

\bibitem{dai2019chance}
S.~Dai, S.~Schaffert, A.~Jasour, A.~Hofmann, and B.~Williams, ``Chance
  constrained motion planning for high-dimensional robots,'' in \emph{2019
  International Conference on Robotics and Automation (ICRA)}.\hskip 1em plus
  0.5em minus 0.4em\relax IEEE, 2019, pp. 8805--8811.

\end{thebibliography}

\end{document}